\documentclass[lettersize,journal]{IEEEtran}
\usepackage{amsmath,amsfonts}
\usepackage{algorithmic}
\usepackage{algorithm}
\usepackage{array}
\usepackage[caption=false,font=normalsize,labelfont=sf,textfont=sf]{subfig}
\usepackage{textcomp}
\usepackage{stfloats}
\usepackage{hyperref}
\usepackage{url}
\usepackage{verbatim}
\usepackage{graphicx}
\usepackage{multirow}
\usepackage{cite}
\usepackage{color}
\usepackage{pifont}
\begin{document}

\title{CADFormer: Fine-Grained Cross-modal Alignment and Decoding Transformer for Referring Remote Sensing Image Segmentation}

\author{{Maofu~Liu,~\IEEEmembership{Member,~IEEE,}
        Xin~Jiang,~\IEEEmembership{Member,~IEEE,}
        and~Xiaokang~Zhang,~\IEEEmembership{Senior Member,~IEEE}}
\thanks{This work was supported in part by the National Natural Science Foundation of China under Grant No. 42371374 and the "14th Five Year Plan" Advantageous and Characteristic Discipline Project of Hubei Province (China) under Grant No. 2023D0302. (\textit{Corresponding author: Xiaokang Zhang}).}
\thanks{Maofu Liu and Xin Jiang are with the  School of Computer Science and Technology, Wuhan University of Science and Technology, Wuhan 430065, China. (e-mail: liumaofu@wust.edu.cn; jx@wust.edu.cn).}
\thanks{Xiaokang Zhang is with the School of Information Science and Engineering, Wuhan University of Science and Technology, Wuhan 430081, China (e-mail: natezhangxk@gmail.com).}
}

\markboth{Journal of \LaTeX\ Class Files,~Vol.~14, No.~8, August~2021}%
{Shell \MakeLowercase{\textit{et al.}}: A Sample Article Using IEEEtran.cls for IEEE Journals}


\maketitle
 
\begin{abstract}
Referring Remote Sensing Image Segmentation (RRSIS) is a challenging task, aiming to segment specific target objects in remote sensing (RS) images based on a given language expression. Existing RRSIS methods typically employ coarse-grained unidirectional alignment approaches to obtain multimodal features, and they often overlook the critical role of language features as contextual information during the decoding process. Consequently, these methods exhibit weak object-level correspondence between visual and language features, leading to incomplete or erroneous predicted masks, especially when handling complex expressions and intricate RS image scenes. To address these challenges, we propose a fine-grained cross-modal alignment and decoding Transformer, CADFormer, for RRSIS. Specifically, we design a semantic mutual guidance alignment module (SMGAM) to achieve both vision-to-language and language-to-vision alignment, enabling comprehensive integration of visual and textual features for fine-grained cross-modal alignment. Furthermore, a textual-enhanced cross-modal decoder (TCMD) is introduced to incorporate language features during decoding, using refined textual information as context to enhance the relationship between cross-modal features. To thoroughly evaluate the performance of CADFormer, especially for inconspicuous targets in complex scenes, we constructed a new RRSIS dataset, called RRSIS-HR, which includes larger high-resolution RS image patches and semantically richer language expressions. Extensive experiments on the RRSIS-HR dataset and the popular RRSIS-D dataset demonstrate the effectiveness and superiority of CADFormer. Datasets and source codes will be available at \href{https://github.com/zxk688}{https://github.com/zxk688}.

\end{abstract}


\section{Introduction}
\IEEEPARstart{I}{n} recent years, with the rapid development of remote sensing (RS) and deep learning, the combination of remote sensing and deep learning has become a popular research topic. Deep learning has made significant progress in various remote sensing tasks, including RS image captioning \cite{cheng2022nwpu}, RS visual question answering \cite{zheng2021mutual}, RS semantic segmentation \cite{10721444}, RS visual grounding \cite{li2024language}, and so on. Despite these advancements, the task of referring remote sensing image segmentation (RRSIS) remains an area of exploration. It combines computer vision and natural language processing, aiming to segment the target objects described by a natural language expression in RS images. Compared to traditional RS semantic segmentation and RS instance segmentation \cite{chen2024rsprompter}, RRSIS is more flexible, allowing users to extract specific target objects of interest from images based on their needs. It holds great potential in various fields such as land use categorization \cite{zhu2012assessment}, disaster response \cite{zhang2023cross}, military intelligence generation \cite{zhang2022progress}, environmental monitoring \cite{yuan2020deep}, and agricultural production \cite{weiss2020remote}.

\begin{figure}[!t]
\centering
\includegraphics[width=1\linewidth]{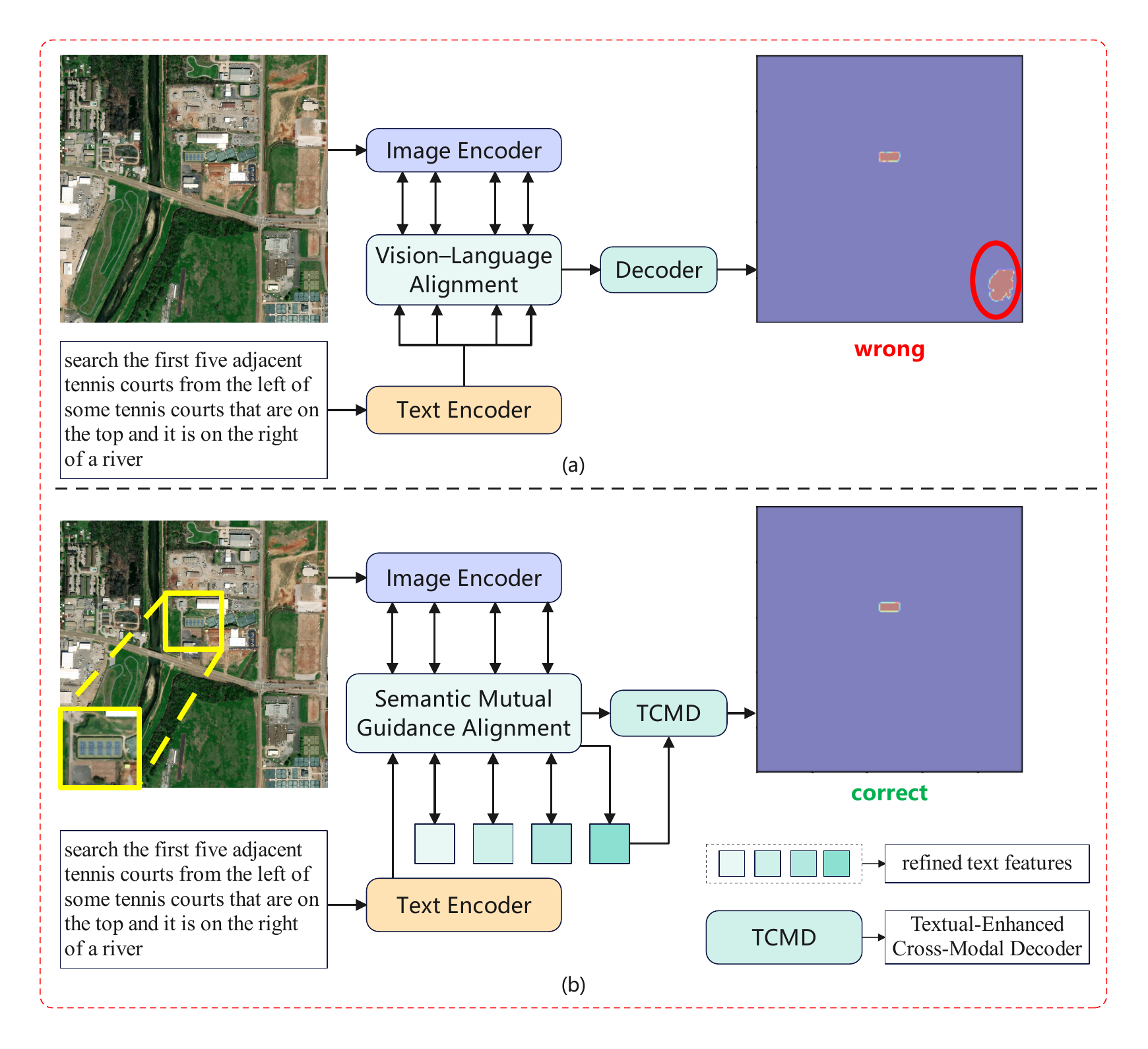}
\caption{Motivation of the proposed approach. The input RS image-text pair is from our proposed RRSIS-HR dataset. (a) Existing RRSIS methods use coarse-grained unidirectional alignment from vision to language and simple standard decoder. (b) Our proposed CADFormer uses semantic mutual guidance alignment and textual-enhanced cross-modal decoder.}
\label{figure_1}
\end{figure}

\IEEEpubidadjcol
Since RRSIS is an emerging task in the field of RS, there has been relatively little exploration in this area. Yuan et al. \cite{yuan2024rrsis} first introduced the concept of RRSIS and proposed the first RRSIS dataset, RefSegRS, along with a language-guided cross-scale enhancement module (LGCE), which aims to improve segmentation performance on small and sparsely distributed objects. Furthermore, to address the issue of complex spatial scales and orientations in RS image scenes, Liu et al. \cite{liu2024rotated} proposed the rotated multi-scale interaction network (RMSIN) and constructed a large-scale RRSIS dataset, called RRSIS-D, based on the RS visual grounding dataset RSVG \cite{zhan2023rsvg}. The proposed RRSIS-D dataset is a new large-scale benchmark for RRSIS tasks and fully advances the research of RRSIS. However, we observe that the referring target objects in the RS images of the RRSIS-D dataset are quite salient, and the referring text descriptions are relatively simple and brief, as shown in Fig. \ref{figure_2}(b). This may decrease the challenge of the RRSIS task, allowing some RIS methods, which perform well on natural images, to still yield good results on the RRSIS-D dataset, even outperforming some RRSIS methods. This drives us to consider whether RRSIS models can still effectively segment inconspicuous targets from very high-resolution RS images with complex language expressions. Therefore, we built a new RRSIS dataset based on the RSVG-HR dataset \cite{lan2024language}, named RRSIS-HR. The RRSIS-HR dataset consists of seven object categories and contains 2650 image-text-label triplets. Compared to the RRSIS-D dataset, the RS images in RRSIS-HR have higher resolution and cover larger regions with complex backgrounds and less prominent objects. Furthermore, the language expressions in RRSIS-HR are longer, more complex and semantically richer, frequently describing multiple object categories, detailed spatial relationships, and contextual information, as shown in Fig. \ref{figure_2}(b). The increased linguistic and visual complexity poses substantial challenges for RRSIS methods.

Previous RRSIS methods \cite{yuan2024rrsis, liu2024rotated} follow the basic architectural strategy in Fig. \ref{figure_1}(a), which adopts simple pixel-word attention \cite{yang2022lavt} to alignment language and visual features, directly integrating original language features derived from BERT \cite{kenton2019bert} into multi-scale visual features throughout the alignment process. This represents a coarse-grained alignment strategy from vision to language and does not fully utilize language features. Ideally, with pixel-word vision-language alignment, language and visual features should exhibit high feature similarity when referring to the same object \cite{chng2024mask}. However, achieving this alignment is not straightforward because language expressions can be highly complex and diverse. When confronted with high-resolution RS image scenes and longer, semantically richer language expressions, these methods exhibit weak object-level correspondence and often struggle to produce optimal segmentation results. In addition, these methods only consider the alignment and interaction between language and visual features during the multimodal feature fusion process. During the decoding phase, cross-modal features are gradually decoded using a simple segmentation head \cite{yang2022lavt, lei2024exploring} to produce the final prediction. However, the importance of text guidance during the decoding process is often overlooked, which may lead to the loss of crucial fine-grained details and suboptimal segmentation results due to the lack of explicit semantic constraints in refining region boundaries and resolving ambiguities.

To address the challenges mentioned above, we propose CADFormer, a novel RRSIS method from the perspective of fine-grained cross-modal alignment and decoding, as shown in Fig. \ref{figure_1}(b). Specifically, we introduce a semantic mutual guidance alignment module (SMGAM) to enhance semantic relevance between cross-modal features. This module performs both language-guided vision alignment and vision-guided language alignment, fully integrating visual and language features from the perspective of mutual guidance alignment. It gradually generates refined multi-scale visual features guided by language and refined language features guided by vision, which serve as input for the subsequent alignment stage and decoding process. This approach differs from previous methods \cite{yuan2024rrsis, liu2024rotated, lei2024exploring} that relied solely on semantic features derived from BERT \cite{kenton2019bert} throughout the alignment process. By fully utilizing language features, we achieve strong object-level correspondence between visual and language features for fine-grained cross-modal alignment. In addition, we design a textual-enhanced cross-modal decoder (TCMD), which accepts the refined multi-scale visual features and language features as input and leverages a Transformer decoder for further processing. The refined language features are used as contextual information to guide the model in processing the refined multi-scale visual features, thus enhancing the interaction between cross-modal features and ensuring more accurate predicted segmentation masks.
In summary, the main contributions of this paper are as follows.
\begin{enumerate}
\item{We propose a novel RRSIS method named CADFormer for handling complex RS image scenes with semantically richer language expressions. Specifically, the SMGAM enhances the semantic relevance between visual and language features by modeling their mutual dependencies and achieves fine-grained cross-modal alignment. Additionally, the TCMD leverages refined language features as contextual guidance for decoding and segmentation, resulting in more accurate predictions.}
\item{We construct a new RRSIS benchmark, RRSIS-HR, which contains high-resolution RS images with fine-grained language expressions, posing challenges for RRSIS methods in handling complex scenes.}
\item{We conduct extensive experiments on the RRSIS-HR and RRSIS-D datasets. The experimental results show that our proposed method, CADFormer, outperforms the majority of existing RRSIS methods, demonstrating the effectiveness of CADFormer and its superiority in handling high-resolution RS image scenes with fine-grained language expressions.}
\end{enumerate}


\section{Related work}
\subsection{Referring Image Segmentation}
Referring image segmentation (RIS) aims to segment specific target objects in images based on natural language expressions, making it a typical multimodal task that has gained increasing attention. Early RIS research \cite{hu2016segmentation} focused on using convolutional neural networks (CNNs) \cite{krizhevsky2012imagenet} and long short-term memory (LSTM) \cite{hochreiter1997long} networks to extract visual and language features, respectively, and then simply fused these features through concatenation to generate the final predictions. Subsequent work improved this process by using recurrent neural networks (RNNs) \cite{liu2017recurrent,li2018referring}, or dynamic networks \cite{han2021dynamic, han2022latency, han2023latency} to progressively refine segmentation masks. The emergence of Transformer \cite{vaswani2017attention} architectures and attention mechanisms revolutionized RIS methods by providing significant fusion capabilities for multimodal integration. CSMA \cite{ye2019cross} effectively captures long-term dependencies between language and visual features through cross-modal self-attention. BRINet \cite{hu2020bi} utilized the bidirectional cross-modal attention module to learn cross-modal relationships between language and visual features. VLT \cite{ding2021vision} employed a cross-attention module to generate query vectors by comprehensively understanding multimodal features, which were then passed through a Transformer decoder to query the given image. To promote cross-modal integration, LAVT \cite{yang2022lavt} introduced a language-aware attention mechanism into the image encoding process, assisting early fusion of cross-modal features and improving segmentation accuracy. Recently, VPD \cite{zhao2023unleashing} explored semantic information in diffusion models \cite{song2020score} for RIS, while RefSegformer \cite{wu2024towards}, ReLA \cite{liu2023gres}, and DMMI \cite{hu2023beyond} focused on improving model robustness. They proposed the generalized RIS task, which differs from conventional RIS where one referring expression corresponds to one target object. In generalized RIS task, a referring expression can refer to an arbitrary number of target objects, including multiple targets or even no target at all. However, unlike natural images where isolated and prominent subjects dominate, the expansive coverage of RS images inevitably captures densely clustered small-scale targets with multi-scale spatial distributions, which are frequently obscured by cluttered backgrounds. These inherent characteristics not only amplify the technical challenges for precise localization but also limit the generalization capability of existing RIS methods in achieving satisfactory performance.

\begin{figure*}[!t]
\centering
\includegraphics[width=1\textwidth]{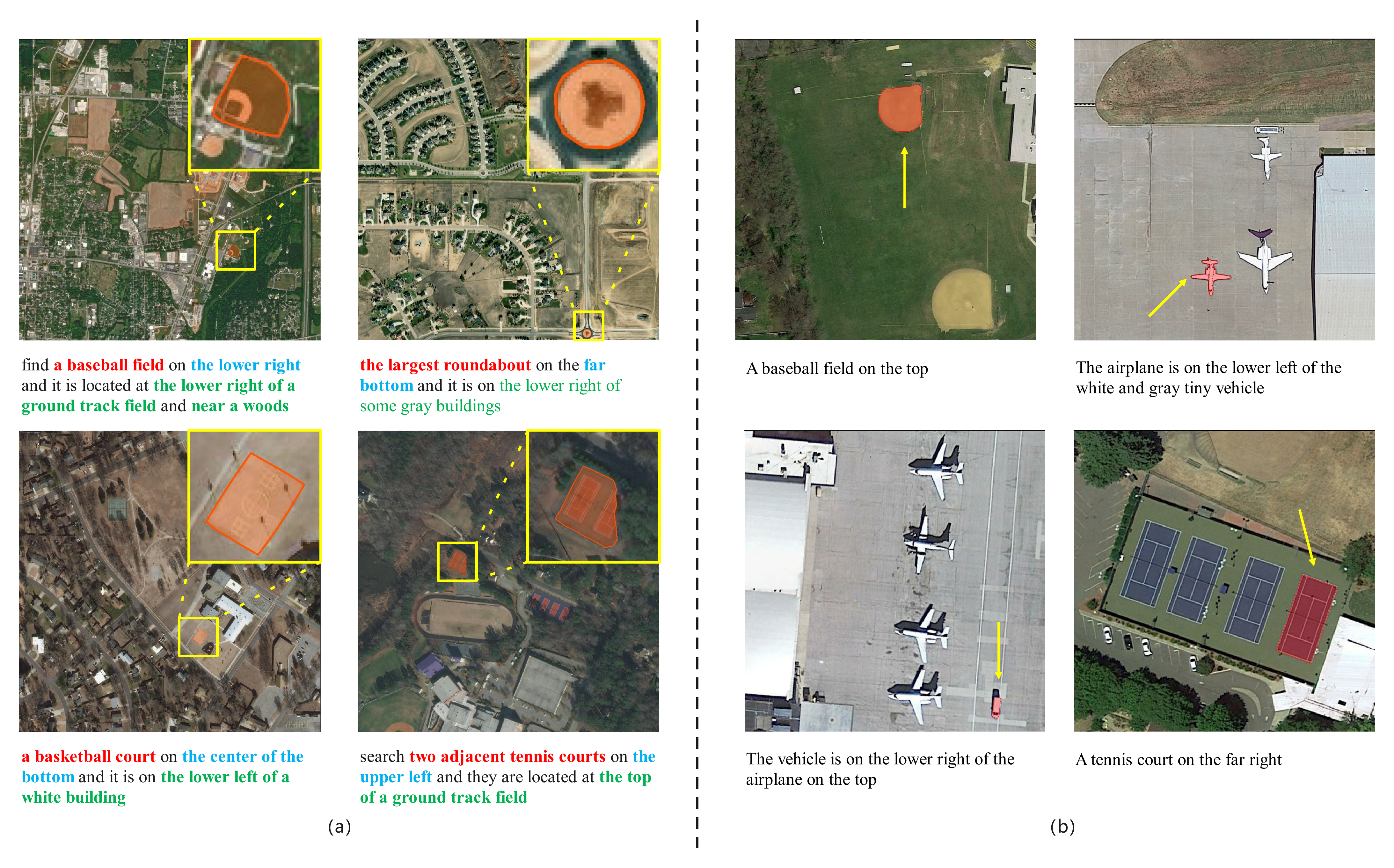}
\caption{Typical examples of our proposed RRSIS-HR datasets and public RRSIS-D datasets. (a) RRSIS-HR dataset. The red, blue, and green fonts in the language expressions represent categories, absolute positions, and relative position relationships, respectively. (b) RRSIS-D dataset.}
\label{figure_2}
\end{figure*}

\subsection{Referring Remote Sensing Image Segmentation}
In recent years, referring image segmentation (RIS) tasks in the domain of remote sensing images have attracted significant attention from researchers. Research in this field is still in its early stages and remains relatively scarce. Yuan et al. \cite{yuan2024rrsis} first introduced the RIS task in the remote sensing domain. They proposed the first RRSIS dataset, RefSegRS, which consists of 4,420 image-language-label triplets. Additionally, based on the LAVT \cite{yang2022lavt} framework, they introduced a new language-guided cross-scale enhancement (LGCE) module. This module utilizes language features to adaptively enhance multi-scale visual features by integrating both deep and shallow features, thus facilitating the detection of small and scattered ground objects. Liu et al. \cite{liu2024rotated} introduced a rotated multi-scale interaction network (RMSIN) to address the challenges posed by varying scales and orientations in remote sensing images. This network, also built on the LAVT framework, effectively explores both intra-scale and cross-scale image-text interactions. To fully evaluate the performance of RMSIN, they introduced a new benchmark dataset, RRSIS-D, which contains 17402 image-text-label triplets, significantly advancing the research and development of RIS. Pan et al. \cite{pan2024rethinking} observed that existing RRSIS methods typically rely on implicit optimization paradigms, which overlook inherent domain gaps and produce class-agnostic predictions. In response, they introduced a dual alignment network (DANet) featuring an explicit alignment strategy to reduce domain differences by narrowing inter-domain affinity distributions, and a reliable proxy alignment module to strengthen multimodal perception and reduce misleading noise interference. Lei et al. \cite{lei2024exploring} explored a fine-grained image-text alignment method and proposed the FIANet network. This network treats the original referring expressions as contextual text, which is further decoupled into object-specific text and spatial location text. Subsequently, the fine-grained image-text alignment module is utilized to integrate visual features with the decoupled textual features, thereby obtaining discriminative multimodal representations. However, these methods uniformly employ the original semantic features extracted by BERT \cite{kenton2019bert} to interact with multi-scale visual features throughout the alignment and fusion process, and some approaches \cite{yuan2024rrsis} \cite{liu2024rotated} omit textual features during the decoding phase. In contrast, our CADFormer method incorporates cross-modal alignment across four stages. Throughout this process, we progressively obtain refined language features and refined visual features based on semantic mutual guidance, which serve as input for subsequent stages. We fully leverage language features, thereby establishing a robust object-level correspondence between visual and language features.

\section{New Benchmark}
\subsection{RRSIS-HR Dataset Construction}
We introduce a new referring remote sensing image segmentation dataset, named RRSIS-HR. Inspired by SAM \cite{kirillov2023segment} and RMISN \cite{liu2024rotated}, we adopt a semi-automatic method, using bounding boxes and SAM to generate pixel-level segmentation masks, significantly reducing the cost of manual annotation. Specifically, we follow the steps below to construct the RRSIS-HR dataset.
\begin{itemize}
\item{Step 1: We collect remote sensing images, referring text descriptions, and corresponding visual grounding bounding boxes from the RSVG-HR \cite{lan2024language} dataset. By leveraging the bounding boxes provided by the RSVG-HR dataset and the bounding box hinting function of SAM, we obtain preliminary pixel-level masks for all referring target objects in the dataset. However, due to the significant domain gap between natural images and remote sensing images, SAM may generate unsatisfactory segmentation masks when applied to partial images, requiring refinement and optimization.}
\item{Step 2: To obtain more accurate fine-grained pixel-level segmentation masks, we optimize the masks generated by SAM in Step 1 through manual verification. First, three annotation experts in the field of remote sensing, drawing on their expertise, developed a set of annotation standards. Following the standards, they carefully examined the masks generated by SAM, identifying and filtering out problematic masks. Subsequently, each annotator used the image segmentation semi-automatic annotation tool \cite{yatenglghorffmanwang} to optimize the problematic masks. This process includes refining boundaries, adjusting sizes, correcting errors, and resolving occlusion issues. When initial corrections are uncertain, a consensus process is triggered. This process involves independent review and cross-checking by three annotators and discussion among multiple experts to make a final decision. Through this human-computer collaborative semi-automatic annotation method, we achieve high-precision mask annotations while significantly reducing the cost of manual annotation.}
\item{Step 3: Finally, to improve the compatibility of RRSIS-HR with various referring image segmentation models, we convert the annotation format to the RefCOCO dataset \cite{lin2014microsoft} format for later use.}
\end{itemize}

\subsection{Data Analysis}
Through the semi-automatic annotation method, we have successfully constructed the RRSIS-HR dataset, which consists of 2650 image-text-mask triplets and includes 7 object categories. Each RS image has a size of 1024×1024 pixels, containing varying scales and details, and covers an area ranging from 0.06 $km^{2}$ to 25 $km^{2}$ \cite{sun2022visual}. The average length of the language descriptions is 19.6 words, with a minimum of 6 words and a maximum of 41 words. Specifically, a language expression contains one or more object categories, which requires RRSIS models to accurately identify target objects from more object categories. Fig. \ref{figure_2} shows some visual examples of RRSIS-HR and RRSIS-D datasets. Compared to the RRSIS-D dataset, although the RRSIS-HR dataset is not large in scale, it contains higher-resolution remote sensing images, with longer and more complex language expressions, making it a challenging dataset that includes complex RS scenes for RRSIS methods.

\section{Methodology}

\begin{figure*}[!t]
\centering
\includegraphics[width=1\linewidth]{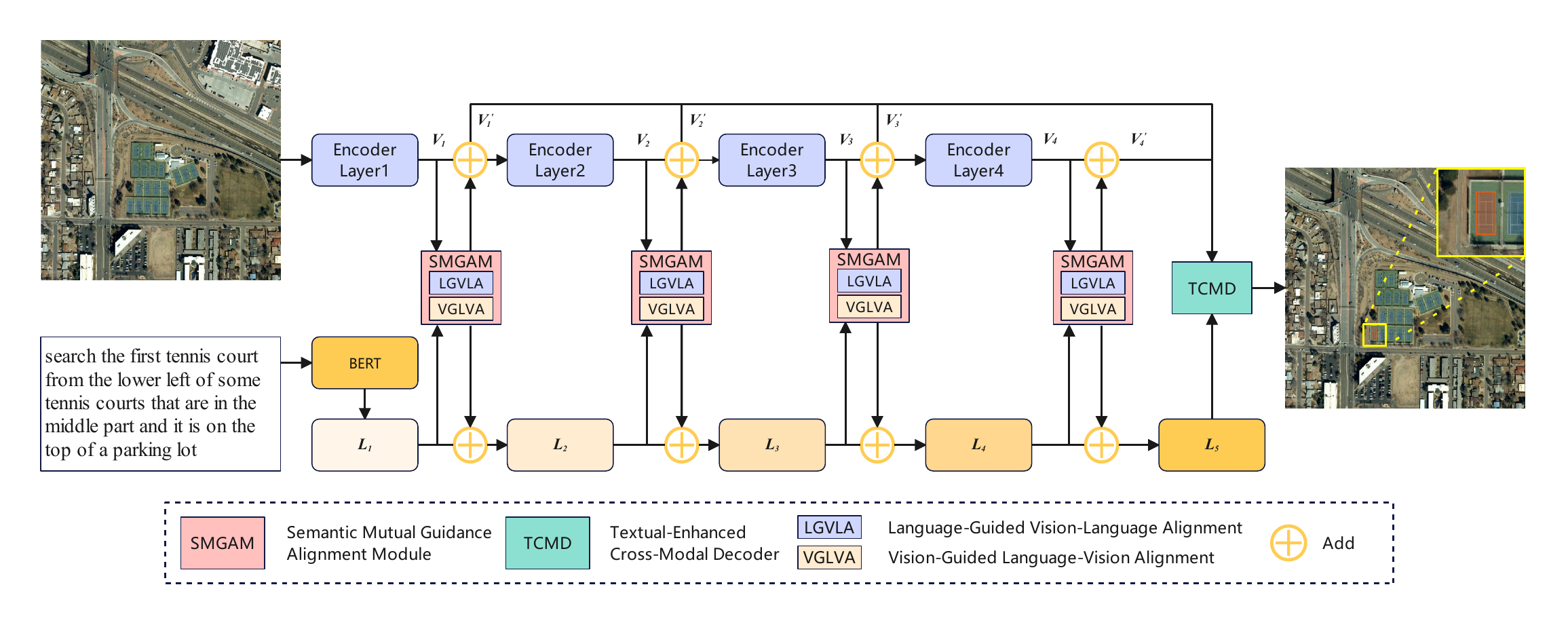}
\caption{Overview of our proposed CADFormer framework. The model first aligns multi-scale visual features and text features progressively through the semantic mutual guidance alignment module (SMGAM). Then, the refined text features are used as contextual information to query the refined multi-scale visual features in the textual-enhanced cross-modal decoder (TCMD), retrieving and aggregating target object information to generate the prediction results.}
\label{figure_3}
\end{figure*}

\subsection{Overview}
The overview of our proposed CADFormer is illustrated in Fig. \ref{figure_3}. It mainly consists of three parts: the image and text encoders, the semantic mutual guidance alignment module (SMGAM), and the textual-enhanced cross-modal decoder (TCMD). Given an image $I \in R^{H \times W \times 3}$ and a language expression $E \in\left\{e_{i}\right\}, i=\{0,1 \ldots N\}$, where $H$ and $W$ denote the height and width of the input image, and $N$ represents the length of the language expression. We utilize the Swin Transformer \cite{liu2021swin} as the visual backbone network to extract multi-scale visual features from the input image. The visual features at each stage are denoted as $V_{i} \in R^{H_{i} \times W_{i} \times C_{i}}, i \in\{1,2,3,4\}$, where $H_{i} $, $W_{i} $ and $C_{i} $ represent the number of height, width and channel of the feature map from the $i$-th stage, respectively. In addition, we use BERT \cite{kenton2019bert} as the text encoder to extract language features represented as $L_{1} \in R^{N\times C_{t} } $, where $N$ and $C_{t} $ denote the length of the sentence and the channel number, respectively. The language features from different stages and the multi-scale visual features are progressively fed into the semantic mutual guidance alignment module. This module aligns cross-modal semantics and generates enhanced visual and linguistic representations. The enhanced visual and linguistic representations are then fed into the textual-enhanced cross-modal decoder to predict pixel-level segmentation masks. We will introduce each module in detail below.

\subsection{Semantic Mutual Guidance Alignment Module}
In prior works \cite{yuan2024rrsis,liu2024rotated,yang2022lavt}, cross-modal alignment was limited to a coarse-grained unidirectional alignment from vision to language. The language features fused at each stage were directly derived from the initial sentence features extracted by BERT \cite{kenton2019bert}. We consider this a coarse-grained alignment strategy that does not fully exploit the potential of language features. When the language expressions increase in length and complexity, existing approaches often encounter challenges in accurately distinguishing and localizing target objects across multiple categories. To address the challenge, we propose the semantic mutual guidance alignment module (SMGAM), which comprises two submodules, namely the language-guided vision-language alignment (LGVLA) submodule and the vision-guided language-vision alignment (VGLVA) submodule, as shown in Fig. \ref{figure_4}. Both submodules take multi-scale visual features $V_{i} $ and stage-specific language features $L_{i} $  as input. The LGVLA submodule produces refined visual features $V_{i}^{'} $, while the VGLVA submodule produces refined language features $L_{i}^{'} $. Through a four-stage semantic mutual guidance alignment process, progressively refined visual and language features are obtained, achieving fine-grained cross-modal alignment. We will describe these two submodules in detail as follows.

\subsubsection{Language-Guided Vision-Language Alignment} During the stage $i$, the LGVLA submodule takes the language features $L_{i}$ and visual features $V_{i}$ as input, as shown in Fig. \ref{figure_4}(a). To better adapt to subsequent alignment tasks, visual features $V_{i}$ are first passed through a projection layer, where they are mapped into a new feature space. The projection layer consists of a 1×1 convolutional layer followed by a GELU activation function, denoted as $\operatorname{Proj}\left ( \cdot  \right ) $. The process can be formulated as: $V_{i} \longleftarrow \operatorname{Proj}\left ( V_{i}  \right ).$ Next, the language features interact with the visual features through a multi-head attention \cite{vaswani2017attention} layer. For cross-modal interaction at each stage, the multi-head attention layer performs cross-attention to obtain enhanced visual representations, where the visual features act as the query, and the language features serve as the key and value. Specifically, the attention layer initially calculates the similarity between the visual features and the language features through a scaled dot-product operation, which aligns each visual element with each language element. This computation results in a cross-modal similarity matrix $M_{vl}^{i}$ that quantifies the relevance and interaction strength between the two modalities. The formulation is as follows:
\begin{equation}
    M_{v l}^{i}=\operatorname{Softmax}\left(\frac{V_{i} W_{q}^{i} \cdot\left(L_{i} W_{k}^{i}\right)^{T}}{\sqrt{C_{i}}}\right)
\end{equation}
where $W_{q}^{i}$ and $W_{k}^{i}$ are the linear projection matrices. Subsequently, we utilize the similarity matrix $M_{vl}^{i}$ to integrate object-relevant details from the visual features into the language features and then multiply the result with the projected visual features, yielding language-guided visual features $A_{v l}^{i}$. The resulting features $A_{v l}^{i}$ are further processed through another projection layer followed by a language gate \cite{yang2022lavt} to produce the refined visual features. The process is specifically described as:
\begin{equation}
    A_{v l}^{i}=M_{v l}^{i} L_{i} W_{v}^{i} \otimes V_{i}
\end{equation}
\begin{equation}
    A_{vl}^{i} \longleftarrow \operatorname{Proj}\left ( A_{vl}^{i} \right ) 
\end{equation}
\begin{equation} 
    V_{i}^{'} = \operatorname{Gate}\left ( A_{vl}^{i}  \right ) 
\end{equation}
\begin{equation}
    \operatorname{Gate}\left(A_{vl}^{i}\right)=\operatorname{MLP}\left(A_{vl}^{i}\right) \otimes A_{vl}^{i}
\end{equation}
where $W_{v}^{i}$ is the linear projection matrice, $\otimes $ denotes element-wise multiplication and $\operatorname{Proj}\left ( \cdot  \right ) $ represents the projection layer, which consists of a 1×1 convolutional layer followed by a GELU  activation function. $\operatorname{Gate}\left ( \cdot  \right ) $ refers to a language gate and MLP is a two-layer perceptron. The first layer is a linear layer followed by a ReLU activation function, while the second layer is a linear layer followed by a Tanh \cite{fan2000extended} activation function.
Subsequently, the refined visual features $V_{i}^{'}$ generated by the LGVLA submodule are merged with the original input features $V_{i}$. After passing through the next stage of the Swin Transformer layer, they are transformed into visual features for the next stage. The process can be described as follows:
\begin{equation}
    V_{i+1}=\operatorname{SwinStage}_{i+1} \left ( V_{i}^{'} + V_{i}  \right )  
\end{equation}
where $\operatorname{SwinStage}_{i}$ denotes the i-th stage of the Swin Transformer \cite{liu2021swin}, which primarily consists of downsampling operations and MLP layers.

\subsubsection{Vision-Guided Language-Vision Alignment} 
\begin{figure}[!t]
\centering
\includegraphics[width=1\linewidth]{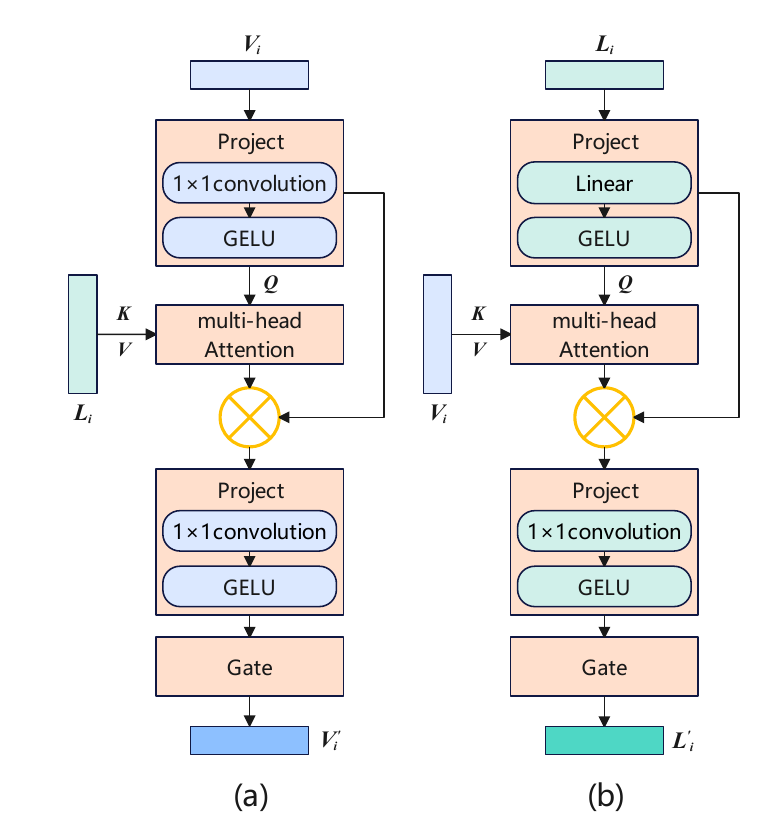}
\caption{Illustration of the proposed SMGAM. Project denotes the projection layer. Gate denotes the gate network. (a) Language-Guided Vision-Language Alignment Submodule. (b) Vision-Guided Language-Vision Alignment Submodule.}
\label{figure_4}
\end{figure}

Similar to the LGVLA submodule, the VGLVA submodule also takes language features and visual features from different stages as input and progressively performs cross-modal interactions between the two modalities. However, unlike the LGVLA submodule, the VGLVA submodule performs a different process and focuses on enhancing the language features through a series of iterative refinements guided by the visual context, as shown in Fig. \ref{figure_4}(b). Specifically, the module employs the multi-head attention \cite{vaswani2017attention} mechanism to iteratively align and refine the language features based on visual information. The output refined language features iteratively interact with the visual features of subsequent stages and this process continues until the refined language features have fully engaged with the multi-scale visual features across all stages. During the $i$-th stage, the language features first pass through a projection layer consisting of a linear layer followed by a GELU activation function, denoted as $\operatorname{Proj}\left ( \cdot  \right ) $. The process is described as follows:
$L_{i} \longleftarrow \operatorname{Proj}\left ( L_{i}  \right ).$    
Subsequently, the multi-head attention layer performs cross-attention to complete the cross-modal interaction between language and visual features, where the language features serve as the query, and the visual features act as the key and value. Specifically, the attention layer initially calculates the similarity between each language element and each visual element through a scaled dot-product operation to obtain the cross-modal similarity matrix $M_{lv}^{i}$ as follows:
\begin{equation}
    M_{l v}^{i}=\operatorname{Softmax}\left(\frac{L_{i} W_{q}^{i} \cdot\left(V_{i} W_{k}^{i}\right)^{T}}{\sqrt{C_{i}}}\right)
\end{equation}
where $W_{q}^{i} $ and $W_{k}^{i} $ are the linear projection matrices. Then, we utilize this similarity matrix $M_{lv}^{i}$ to integrate object-relevant information from the language features into the visual features and multiply the result with the projected language features to obtain vision-guided language features. The resulting output is then passed through a projection layer, followed by a gate network similar to the language gate \cite{yang2022lavt}, to produce the refined language features. This gate network is still denoted as $\operatorname{Gate}\left ( \cdot  \right ) $. The process is described as follows:
\begin{equation}
    A_{l v}^{i}=M_{l v}^{i} V_{i} W_{v}^{i} \otimes L_{i}
\end{equation}
\begin{equation}
    A_{lv}^{i} \longleftarrow \operatorname{Proj}\left ( A_{lv}^{i} \right ) 
\end{equation}
\begin{equation} 
    L_{i}^{'} = \operatorname{Gate}\left ( A_{lv}^{i}  \right ) 
\end{equation}
\begin{equation}
    \operatorname{Gate}\left(A_{lv}^{i}\right)=\operatorname{MLP}\left(A_{lv}^{i}\right) \otimes A_{lv}^{i}
\end{equation}
where $W_{v}^{i} $ is the linear projection matrice, $\otimes $ denotes element-wise multiplication. $\operatorname{Gate}\left ( \cdot  \right ) $ denotes as a gate network similar to the language gate \cite{yang2022lavt}, also consisting of MLP \cite{liu2021swin} and multiplication operation.
After that, the refined language features generated by the VGLVA submodule are merged with the input features $L_{i} $ and transformed into the language features for the next stage, serving as the input for the subsequent SMGAM stage. The process is described as follows:
\begin{equation}
    L_{i+1}=L_{i}+L_{i}^{'}
\end{equation}
The refined language features obtained at the final stage are denoted as $L_{5}$.

\subsection{Textual-Enhanced Cross-modal Decoder}
Previous works \cite{yuan2024rrsis, liu2024rotated, yang2022lavt, lei2024exploring} used only refined multi-scale visual features as input during mask prediction, without fully leveraging the language features. These methods typically perform cross-modal interaction only before decoding, limiting their ability to further utilize language information during the decoding process. As a result, fine-grained details crucial for accurate segmentation may be lost. Furthermore, the lack of deep interaction between the semantic details in the language features and the visual features hinders the ability of the model to effectively distinguish between different categories or precisely segment specific target objects within the same category.

\begin{figure*}[!t]
\centering
\includegraphics[width=0.9\textwidth]{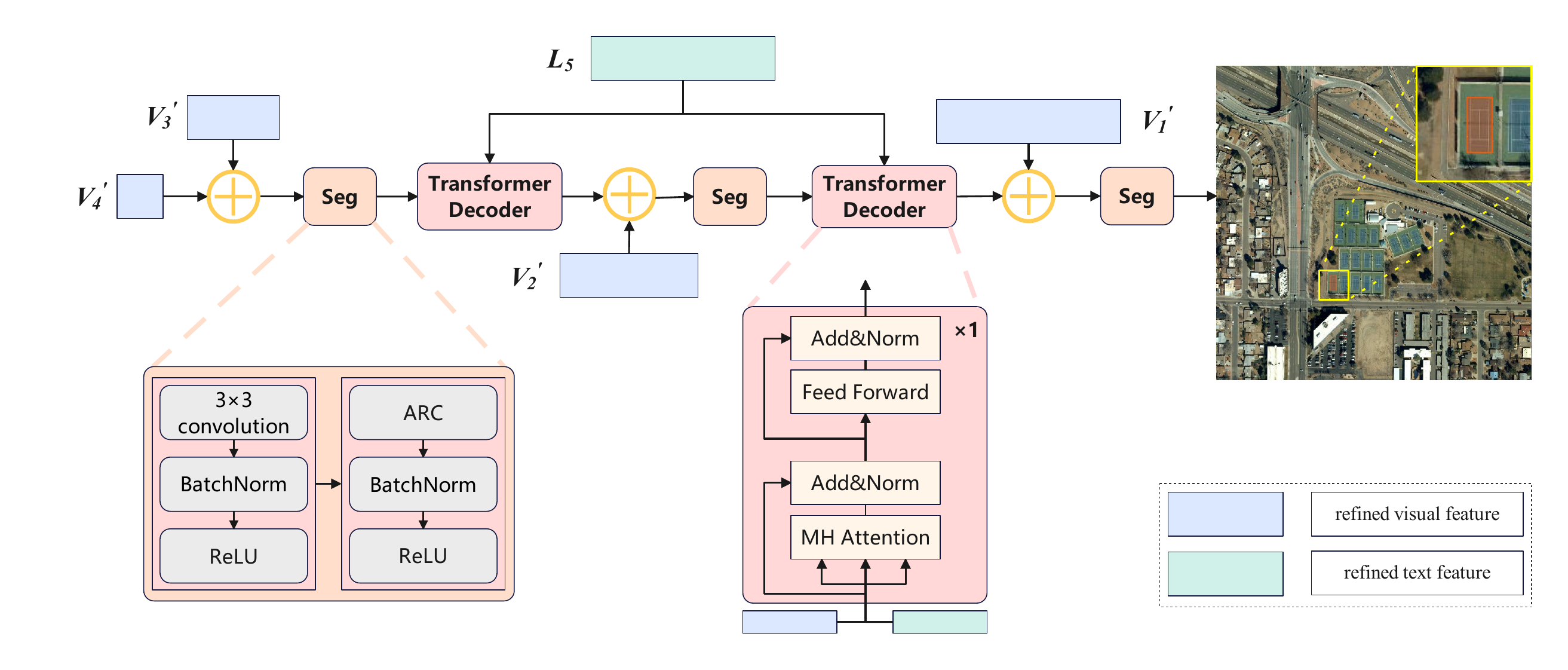}
\caption{Illustration of the proposed TCMD. MH Attention denotes the multi-head attention layer. ARC denotes the adaptive rotated convolution.}
\label{figure_5}
\end{figure*}

In contrast, our textual-enhanced cross-modal decoder (TCMD) utilizes refined multi-scale visual features and refined language features $L_{5}$ as input for mask prediction, as shown in Fig. \ref{figure_5}. The refined language features serve as contextual information to gradually guide the feature decoding process, retrieving and aggregating target object information from the refined visual features. This enables the model to capture the fine-grained relationships between modalities more accurately, thereby enhancing its understanding of complex scenes and improving the quality of mask prediction. To further enhance the interaction between cross-modal features, we incorporate the Transformer decoder \cite{vaswani2017attention} into this process. Specifically, our decoding process follows a top-down approach, integrating refined multi-scale visual features with refined language features. The overall process can be described as follows:
\begin{equation}
    Y_{4}=V_{4}^{'}
\end{equation}
\begin{equation}
    Y_{i}=\operatorname{Transformer}\left(\operatorname{Seg}\left(\left[Y_{i+1} ; V_{i}^{\prime}\right]\right), L_{5}\right), i=\{2,3\}
\end{equation}
\begin{equation}
    Y_{1}=\operatorname{Seg}\left(\left[Y_{2} ; V_{1}^{\prime}\right]\right)
\end{equation}
where $\left [ ; \right ] $ represents the concatenation operation along the channel dimension. $\operatorname{Seg}\left ( \cdot  \right ) $  consists of two 3×3 convolutional layers, batch normalization, and ReLU activation functions to enhance the non-linearity of the segmentation feature space. Additionally, one of the 3×3 convolutional layers is replaced by an adaptive rotated convolution layer \cite{liu2024rotated} to leverage directional information in the feature space, thereby eliminating redundancy and improving the accuracy of boundary details. $\operatorname{Transformer}\left ( \cdot  \right ) $ represents the Transformer decoder layer. The final feature map is projected into two class score maps using a 1×1 convolution. Finally, bilinear interpolation is employed to upsample the results to match the resolution of the input image.

\subsection{Loss Function}
In remote sensing images, the scarcity of target pixels compared to the abundance of background pixels creates a notable class imbalance. This imbalance can lead traditional cross-entropy loss functions to bias the model towards learning background features, ultimately reducing the effectiveness of target region detection. To address this issue, we adopt a combined loss function consisting of cross-entropy loss and Dice loss as our training objective:
\begin{equation}
    \mathcal{L}=\lambda \cdot \mathcal{L}_{\text {cross-entropy }}\left(Y, \hat{Y}\right)+(1-\lambda) \cdot  \mathcal{L}_{\text {dice }}\left(Y, \hat{Y}\right)
\end{equation}
where $\lambda$ is the hyperparameter that balances the two loss functions, set to 0.9. $\hat{Y} $ represents the predicted results, and $Y$ denotes the ground truth.

\begin{table}[!htbp]
\caption{DETAIL OF THE RRSIS-HR DATASET}
    \centering
    \label{table1}
    \begin{tabular}{ccccc}
    \hline
        Category & train & val & test & Total \\ \hline
        Baseball field & 420 & 53 & 52 & 525 \\ 
        Basketball field & 190 & 24 & 24 & 238 \\ 
        Ground track field & 170 & 22 & 21 & 213 \\ 
        Roundabout & 524 & 66 & 66 & 656 \\ 
        Swimming pool & 152 & 19 & 19 & 190 \\ 
        Storage tank & 170 & 22 & 21 & 213 \\ 
        Tennis court & 492 & 62 & 61 & 615 \\ 
        Total & 2118 & 268 & 264 & 2650 \\ \hline
    \end{tabular}
\end{table}

\section{Experiments}
\subsection{Implementation Details}
\subsubsection{Experiment Settings} We use PyTorch to implement our method. Similar to previous methods \cite{yuan2024rrsis, liu2024rotated}, during the experiments, we use the Swin Transformer \cite{liu2021swin} as the visual backbone, pre-trained on ImageNet22K \cite{deng2009imagenet}, and the base BERT model from the HuggingFace’s library \cite{wolf2020transformers} as the text encoder. We employ the AdamW optimizer \cite{loshchilov2017decoupled} with a weight decay of 0.01 and an initial learning rate of 0.00005, with the learning rate decaying according to a polynomial schedule. The batch size is set to 2, and each model is trained for 40 epochs on an NVIDIA GeForce RTX 3090 GPU. During both training and testing phases, all images are resized to 480×480 pixels.
\subsubsection{Evaluation Metrics} Following the prior research \cite{yuan2024rrsis, liu2024rotated, yang2022lavt}, we use mean intersection over union (mIoU), overall intersection over union (oIoU), and Precision@X (Pr@X) as evaluation metrics. Precision@X refers to the percentage of test samples for which the IoU between the predicted result and the ground truth exceeds a threshold X. It is used to evaluate the accuracy at a specific IoU threshold and reflects the method's performance in object localization. The mIoU and oIoU can be formulated as follows:
\begin{equation}
    mIoU=\frac{1}{M} \sum_{t} I_{t} / U_{t}
\end{equation}
\begin{equation}
    oIoU=\sum_{t} I_{t}/\sum_{t} U_{t}
\end{equation}
where $t$ is the index of the image-language-label triplets and $M$ represents the size of the dataset. $I_{t}$ and $U_{t}$ are the intersection and union area of predicted and ground-truth regions.
\subsubsection{Compared Methods}
To evaluate the effectiveness of our proposed CADFormer, we compared it with several state-of-the-art methods of referring image segmentation for both natural images and remote sensing images on the test sets of RRSIS-D and RRSIS-HR. The results of the different methods are shown in Tables \ref{table2} and Table \ref{table3}, respectively. For a fair comparison, we reimplemented some of the state-of-the-art methods, including LAVT \cite{yang2022lavt}, LGCE \cite{yuan2024rrsis}, RMSIN \cite{liu2024rotated} and FIANet \cite{lei2024exploring}, with a total of 40 training epochs for both RRSIS-D and RRSIS-HR. For some earlier released methods, we used the results reported in RMISN \cite{liu2024rotated}.

\begin{table*}[!htbp]
\caption{Comparison With State-of-the-art Methods on the RRSIS-D Test Set. The Best Result is Bold. }
    \centering
    \label{table2}
    \begin{tabular}{lccccccc}
    \hline
        Method & Pr@0.5 & Pr@0.6 & Pr@0.7 & Pr@0.8 & Pr@0.9 & mIoU & oIoU \\ \hline
        RRN \cite{li2018referring} & 51.07 & 42.11 & 32.77 & 21.57 & 6.37 & 45.64 & 66.43 \\ 
        CSMA \cite{ye2019cross} & 55.32 & 46.45 & 37.43 & 25.39 & 8.15 & 48.54 & 69.39 \\ 
        LSCM \cite{hui2020linguistic} & 56.02 & 46.25 & 37.70 & 25.28 & 8.27 & 49.92 & 69.05 \\ 
        CMPC \cite{huang2020referring} & 55.83 & 47.40 & 36.94 & 25.45 & 9.19 & 49.24 & 69.22 \\ 
        BRINet \cite{hu2020bi} & 56.9 & 48.77 & 39.12 & 27.03 & 8.73 & 49.65 & 69.88 \\ 
        CMPC+ \cite{liu2021cross} & 57.65 & 47.51 & 36.97 & 24.33 & 7.78 & 50.24 & 68.64 \\ 
        LAVT \cite{yang2022lavt} & 62.32 & 56.16 & 47.06 & 36.37 & 21.49 & 55.48 & 74.68 \\ 
        LGCE \cite{yuan2024rrsis} & 70.44 & 63.95 & 53.86 & 41.94 & \textbf{24.53} & 61.01 & 76.55 \\ 
        FIANet \cite{lei2024exploring} & 72.16 & 65.81 & 54.67 & 41.60 & 24.48 & 62.58 & 76.48 \\ 
        RMSIN \cite{liu2024rotated} & 72.22 & 65.84 & 55.44 & 42.26 & 24.10 & 62.35 & 76.67 \\ 
        CADFormer (Ours) & \textbf{74.20} & \textbf{67.62} & \textbf{55.59} & \textbf{42.37} & 23.59 & \textbf{63.77} & \textbf{77.26} \\ \hline
    \end{tabular}
\end{table*}

\subsection{Dataset}
We conducted experiments on two datasets, including the publicly available RRSIS-D dataset and the RRSIS-HR dataset constructed by us. Detailed information about these datasets is as follows.

\subsubsection{RRSIS-D} The RRSIS-D dataset is built on the DIOR-RSVG \cite{zhan2023rsvg} dataset and contains 20 object categories. The dataset contains a total of 17402 image-language-label triplets, with 12181 for training, 1740 for validation, and the rest 3481 for testing, which is a large benchmark. The image size in this dataset is 800×800 pixels with spatial resolutions ranging from 0.5m to 30m. The average length of the language expressions is 6.8 words. Some visual examples of the RRSIS-D dataset are shown in Fig. \ref{figure_2}(b).

\subsubsection{RRSIS-HR} The RRSIS-HR dataset contains very high-resolution RS images and longer, semantically richer language expressions. Specifically, the dataset contains 2650 image-language-label triplets and 7 object categories in total. The training set has 2118 triplets, the validation set has 268 triplets, and the rest 264 triplets are in the test set. Each RS image has a size of 1024×1024 pixels, containing varying scales and details, and covers an area ranging from 0.06 $km^{2} $ to 25 $km^{2} $. The average length of the language descriptions is 19.6 words, with a minimum of 6 words and a maximum of 41 words. More detailed information about the  RRSIS-HR dataset can be found in Table \ref{table1}, and some visual examples are shown in Fig. \ref{figure_2}(a).

\begin{figure*}[!htbp]
\centering
\includegraphics[width=0.8\textwidth]{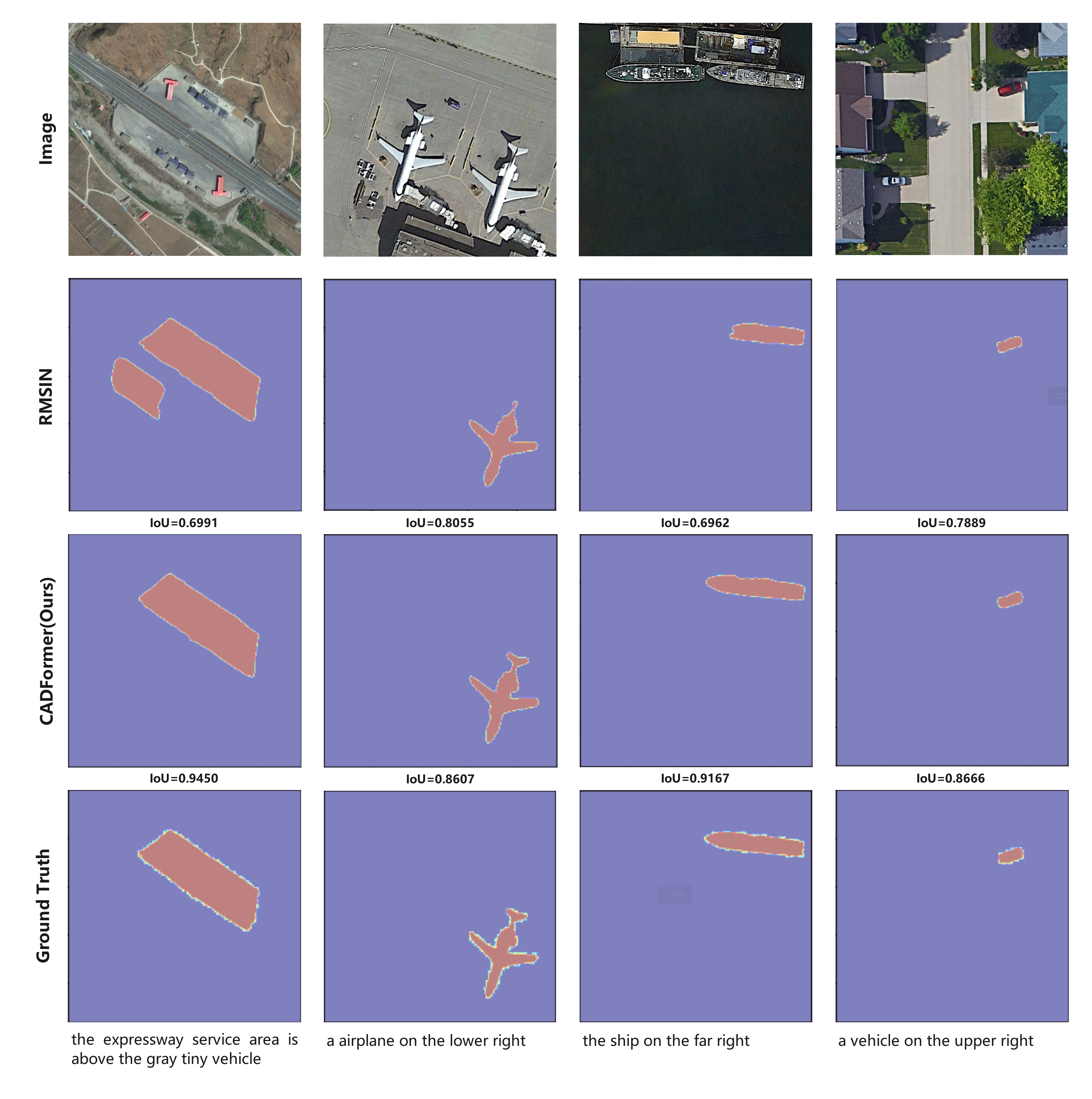}
\caption{Qualitative comparisons of different methods on the RRSIS-D test set. From top to bottom: original image, predictions by RMSIN, predictions by CADFormer, ground truth, and language expressions.}
\label{figure_6}
\end{figure*}

\begin{table*}[!htbp]
\caption{Comparison With State-of-the-art Methods on the RRSIS-HR Test Set. The Best Result is Bold. }
    \centering
    \label{table3}
    \begin{tabular}{lccccccc}
    \hline
        Method & Pr@0.5 & Pr@0.6 & Pr@0.7 & Pr@0.8 & Pr@0.9 & mIoU & oIoU \\ \hline
        LAVT \cite{yang2022lavt} & 23.11 & 20.08 & 13.64 & 5.3 & 0.38 & 22.78 & 27.94 \\ 
        FIANet \cite{lei2024exploring} & 31.06 & 27.65 & 22.35 & 15.15 & 1.89 & 27.13 & 28.89 \\ 
        LGCE \cite{yuan2024rrsis} & 35.98 & 31.06 & 23.86 & 15.15 & 3.79 & 33.48 & 38.20 \\ 
        RMSIN \cite{liu2024rotated} & 50.00 & 46.97 & 39.77 & 29.92 & 6.44 & 43.70 & 45.97 \\ 
        CADFormer (Ours) & \textbf{61.74} & \textbf{57.95} & \textbf{48.86} & \textbf{35.61} & \textbf{13.26} & \textbf{54.71} & \textbf{53.29} \\ \hline
    \end{tabular}
\end{table*}

\subsection{Results and Analysis}

\begin{figure*}[!htbp]
\centering
\includegraphics[width=0.8\textwidth]{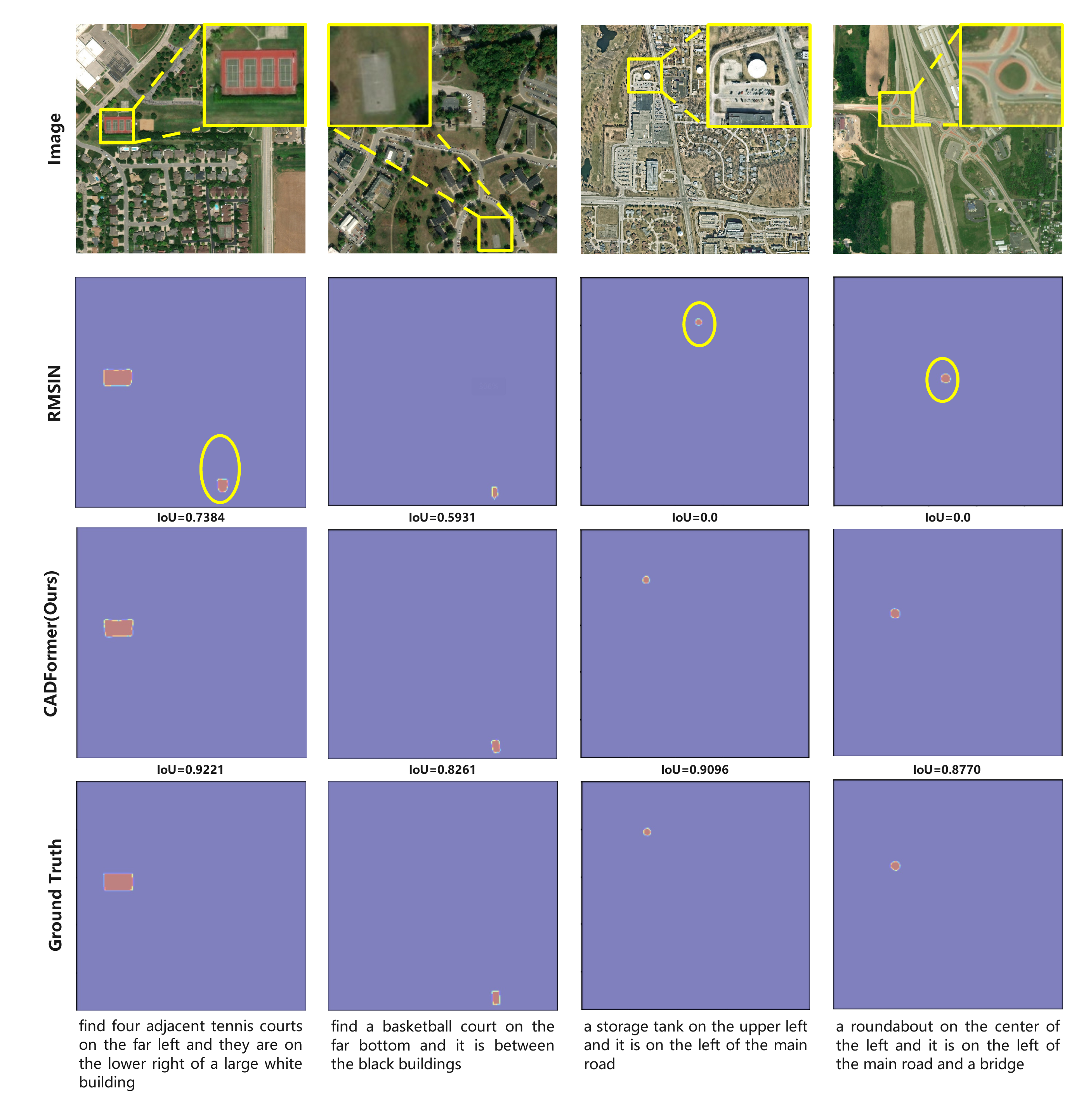}
\caption{Qualitative comparisons of different methods on the RRSIS-HR test set. From top to bottom: original image, predictions by RMSIN, predictions by CADFormer, ground truth, and language expressions. Yellow boxes indicate the approximate location of the target objects, while yellow circles highlight obvious incorrect predictions.}
\label{figure_7}
\end{figure*}

Table \ref{table2} presents the overall results of different methods on the RRSIS-D dataset. It can be observed that our method achieves the best performance across multiple evaluation metrics, including mIoU, oIoU, and precision scores from Pr@0.5 to Pr@0.8. Notably, our method outperforms the second-best RMSIN by 1.42\% in mIoU, 0.59\% in oIoU, and 1.98\% in Pr@0.5, and 1.78\% in Pr@0.6. This result suggests that our method achieves strong segmentation performance at low and medium overlap thresholds. However, its performance at the stricter Pr@0.9 threshold is less competitive, potentially due to the limitations of the model in capturing high-precision details, as it prioritizes overall segmentation accuracy. Fig. \ref{figure_6} shows the visual segmentation results of different methods on the RRSIS-D test set, along with the corresponding IoU scores. It can be observed that, compared to the model RMSIN \cite{liu2024rotated}, our CADFormer exhibits superior segmentation performance across various remote sensing scenes. Specifically, our method produces more accurate pixel-level segmentation masks with higher IoU scores for different ground objects at various scales while significantly reducing misclassification errors.

We further evaluated our proposed method on the RRSIS-HR dataset. Considering the complexity of the dataset, we selected several state-of-the-art RRSIS methods and LAVT \cite{yang2022lavt} as comparison models. The results are shown in Table \ref{table3}. It can be observed that our method achieves the best performance across all metrics. Specifically, our CADFormer outperforms the second-best RMSIN \cite{liu2024rotated} by 10.98\% in Pr@0.6, 11.01\% in mIoU, and 7.32\% in oIoU. Fig. \ref{figure_7} shows the visual segmentation results of different methods on the RRSIS-HR test set, along with the corresponding IoU scores. For better clarity, we marked the approximate locations of the target objects with yellow boxes in the first row of the original images. Additionally, we highlighted the clearly incorrect predicted areas with yellow circles. As can be seen, our method CADFormer demonstrates superior segmentation performance by more accurately segmenting the target objects in complex remote sensing scenes. For instance, in the first in column of Fig. \ref{figure_7}, RMSIN \cite{liu2024rotated} not only predicts the target object but also mistakenly predicts other non-specific category objects, as indicated by the yellow circles. In the third and fourth columns, RMSIN incorrectly predicts the storage tank on the right and the roundabout in the middle, respectively. In contrast, our CADFormer correctly predicts the target objects specified by the language expressions, which aligns with the task requirements of RRSIS. The quantitative and qualitative results on the RRSIS-HR test set indicate that when the resolution of the RS images is very high, and the target objects are hidden in complex backgrounds with more complex language expressions, previous RRSIS methods fail to deliver satisfactory performance. However, our CADFormer can achieve accurate target segmentation, demonstrating its effectiveness and superiority.

\subsection{Ablation Study}
We conducted ablation experiments on the test sets of both the RRSIS-HR and RRSIS-D datasets to verify the effectiveness of the core modules in our method.

\subsubsection{Effectiveness of SMGAM and TCMD} To evaluate the effectiveness of our proposed SMGAM and TCMD, we performed ablation studies on all combinations of SMGAM and TCMD, as illustrated in Table \ref{table4}. The first row presents the experimental results of the model without SMGAM and TCMD, which only reaches 60.05\% and 39.48\% mIoU on the RRSIS-D and RRSIS-HR datasets. The second row shows the results of the model only containing SMGAM. The results show that the introduction of SMGAM improves the mIoU by 2.34\% on the RRSIS-D dataset and by 7.16\% on the RRSIS-HR dataset. In the third row, we add TCMD for the model without SMGAM. The results indicate that the introduction of TCMD leads to improvements in Pr@0.6, mIoU, and oIoU. The fourth row shows the complete model, our CADFormer. Although the value of oIoU is slightly lower by 0.07\% compared to the second row, this complete model outperforms the base model without SMGAM and TCMD in all metrics. Specifically, the value of mIoU improved by 3.72\% and 15.23\% on the two datasets, respectively. These results demonstrate that our proposed SMGAM and TCMD are effective in improving the overall segmentation capability and play a crucial role in handling complex scenes.

\subsubsection{TCMD Module Analysis} 
\begin{figure}[!t]
\centering
\includegraphics[width=\linewidth]{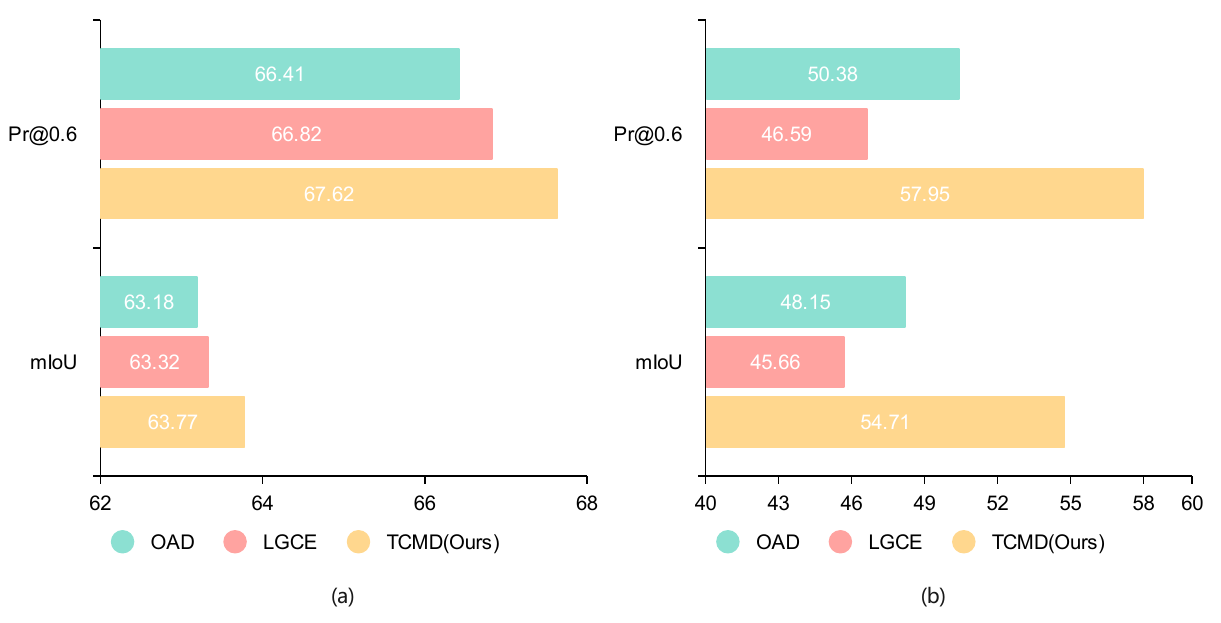}
\caption{TCMD module analysis, where OAD denotes the oriented-aware decoder, and LCGE represents that the model uses the standard decoder in LGCE. (a) Comparison on the RRSIS-D test (b) Comparison on the RRSIS-HR test.}
\label{figure_8}
\end{figure}

To further demonstrate the effectiveness of the proposed TCMD, we compared it with two existing decoder: the oriented-aware decoder (OAD) in RMSIN \cite{liu2024rotated} and the standard segmentation decoder in LGCE \cite{yuan2024rrsis}. We utilize the two decoders to replace the TCMD, respectively, maintaining the integrity of the remaining components within the CADFormer model. Fig. \ref{figure_8} presents the comparative experimental results of using different decoders, where Fig. \ref{figure_8}(a) and Fig. \ref{figure_8}(b) demonstrate the performance of the model on the test sets of the RRSIS-D and RRSIS-HR datasets, respectively. The experimental results indicate that our proposed TCMD method outperforms the other two decoders, which not only validates the importance of incorporating language features at the decoding stage but also confirms the effectiveness of our proposed TCMD module.

\begin{table}[!]
\caption{Ablation Studies on the Smgam and Tcmd}
    \centering
    \label{table4}
    \scalebox{0.9}{
    \begin{tabular}{cc|ccc|ccc}
   \hline
        \multirow{2}*{SMGAM} & \multirow{2}*{TCMD} & \multicolumn{3}{c|}{RRSIS-D} & \multicolumn{3}{c}{RRSIS-HR} \\ 
        & & Pr@0.6 & mIoU & oIoU & Pr@0.6 & mIoU & oIoU \\ \hline
        \ding{55} & \ding{55} & 62.31 & 60.05 & 76.37 & 40.15 & 39.48 & 41.92 \\ 
        \ding{51} & \ding{55} & 65.76 & 62.39 & \textbf{77.33} & 48.86 & 46.64 & 48.96 \\ 
        \ding{55} & \ding{51} & 66.53 & 63.61 & 76.89 & 55.68 & 53.03 & 53.02 \\ 
        \ding{51} & \ding{51} & \textbf{67.62} & \textbf{63.77} & 77.26 & \textbf{57.95} & \textbf{54.71} & \textbf{53.29} \\ \hline
    \end{tabular}}
\end{table}

\section{Conclusion}
In this paper, we propose CADFormer, a novel RRSIS method based on semantic mutual guidance alignment and textual-enhanced cross-modal decoder, which excels at segmenting specific target objects in complex RS scenes. Specifically, SMGAM aims to enhance the semantic correlation between visual and language features and achieve fine-grained cross-modal alignment, generating refined multi-scale visual and refined language features based on semantic mutual guidance. In TCMD, we use the refined language features as contextual information to retrieve and aggregate referential object information from refined multi-scale visual features, achieving precise segmentation. Besides, a new RRSIS dataset based on very high-resolution RS images with longer, semantically richer language expressions is constructed to evaluate the performance of the existing RRSIS methods and our proposed methods in complex RS scene understanding. Experimental results on two RRSIS datasets demonstrate that our CADFormer outperforms the majority of existing RRSIS methods. Additionally, when dealing with complex scenes and language expressions, our method can generate fine-grained segmentation results. From the experiments, we find that semantic mutual guidance alignment facilitates fine-grained cross-modal alignment, while incorporating text features during the decoding process effectively improves segmentation accuracy. Future work could focus on developing a generalized referring image segmentation approach for RS images, which can match multiple targets or no target based on language expressions.

\section*{Acknowledgments}
The authors would like to thank the editors and reviewers for their instructive comments, which helped to improve this article.



 

\bibliographystyle{IEEEtran}
\bibliography{reference}

\vfill

\end{document}